\documentclass[11pt]{article}

\usepackage[T1]{fontenc}
\usepackage[utf8]{inputenc}
\usepackage{newtxtext}
\usepackage{CJKutf8}
\usepackage{setspace}
\usepackage[final]{microtype}
\usepackage{caption}
\pdfmapfile{+fonts/pdftex/notoserifjp.map}

\newcommand{\diffp}[1]{\textcolor{OliveGreen!60!green}{#1}}

\newcommand{\diffm}[1]{\textcolor{red}{#1}}

\newcommand{\diffz}[1]{\textcolor{black}{#1}}

\newcommand{\inlinep}[1]{{\footnotesize\diffp{(#1)}}}
\newcommand{\inlinem}[1]{{\footnotesize\diffm{(#1)}}}
\newcommand{\inlinez}[1]{{\footnotesize\diffz{(#1)}}}

\usepackage[dvipsnames,table]{xcolor}
\usepackage{array}
\usepackage{booktabs}
\usepackage{multirow}
\usepackage{makecell}
\usepackage{tabularx}
\usepackage{colortbl}

\definecolor{hlgreen}{RGB}{205,232,215}

\definecolor{hlpink}{RGB}{255,225,235}

\newcolumntype{Y}{>{\raggedright\arraybackslash}X}

\usepackage[final]{acl}

\usepackage{latexsym}

\usepackage{microtype}

\usepackage{graphicx}

\usepackage[english]{babel}

\title{Beyond Fluent Responses: Finding User Dissatisfaction \\in Human--AI Interactions}

\author{
  \textbf{Yuka Saito}\textsuperscript{1},
  \textbf{Ikumi Numaya}\textsuperscript{1},
  \textbf{Jun Suzuki}\textsuperscript{1,2,3},
  \textbf{Reina Akama}\textsuperscript{1,2,4}
\\
  \textsuperscript{1}Tohoku University,
  \textsuperscript{2}RIKEN,
  \textsuperscript{3}NII LLMC,
  \textsuperscript{4}NINJAL
\\
  {\texttt{is-failab-research@grp.tohoku.ac.jp}}
}

\begin{document}
\begin{CJK}{UTF8}{notoserifjp}
\maketitle

\begin{abstract}

As large language models generate increasingly high-quality responses, recent work on response evaluation has moved beyond response-centered criteria, such as correctness and coherence, toward user-centered criteria, including user preferences and perceived usefulness. 
This shift is fundamentally important for developing technologies that are truly useful to real-world users. 
While conventional response evaluation quantifies the quality of the response itself, users' satisfaction in real conversations also depends on their subjective preferences and perceptions of usefulness, which do not necessarily align with response quality.
This study investigates whether such user-centered signals can be detected from conversational text. 
Specifically, we analyze user utterances with self-reported dissatisfaction labels to examine whether user dissatisfaction in text can be detected.
In English interaction, models using user utterances exceeded the 17.8\% dissatisfaction label prevalence in AUPRC by up to 15 percentage points.
These results suggest that user utterances reveal cues to implicit dissatisfaction.
Furthermore, changes from the immediately preceding user utterance may improve dissatisfaction detection, with anomaly detection methods appearing particularly effective.

\end{abstract}

\section{Introduction}
Recent advances in large language models (LLMs) have enabled increasingly fluent and sophisticated response generation, leading to their growing use in chat-based text interactions, such as ChatGPT\footnote{\url{https://openai.com/chatgpt}}~\cite{Chatterji_2025_HowPeopleUseChatGPT}.
Iterative dialogue between users and LLMs has now become commonplace, and LLMs are increasingly being positioned as communication partners in contexts conventionally dominated by human--human interaction~\cite{yang-etal-2024-human,tanguy-etal-2025-human,10.1145/3764591}.
With the continued expansion of their social use, opportunities for interaction with non-human agents are expected to grow further, raising the standards users expect from such interactions.

\begin{figure}[t]
\centering
\includegraphics[width=0.8\linewidth]{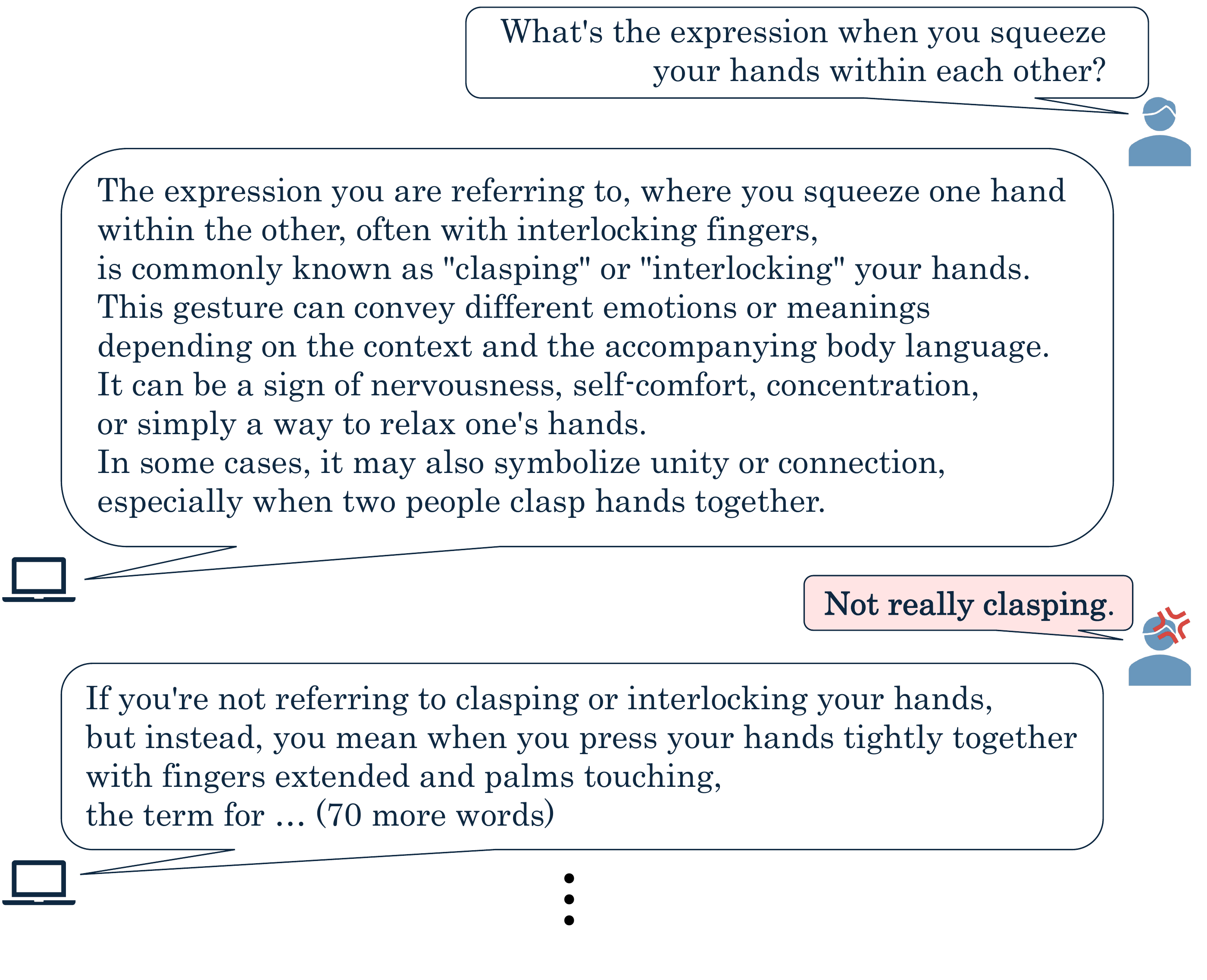}
\caption{Data format for dissatisfaction detection. Given a user utterance after a system response, the task is to detect whether it reflects user dissatisfaction, using self-reported labels as ground truth.}
\vspace{-1em}
\label{fig:overview}
\end{figure}

Traditionally, the evaluation of model-generated responses has focused primarily on the responses themselves, with an emphasis on correctness and coherence~\cite{lowe-etal-2017-towards, 10.5555/3666122.3668142, lin-chen-2023-llm, mendoncca2024benchmarking}.
In recent years,
response-centered evaluations have demonstrated strong performance~\cite{zhang-etal-2020-dialogpt, lin-chen-2023-llm}.
However, users' evaluations of system responses are shaped not only by response qualities, but also by subjective factors such as user preferences~\cite{10.3115/1118053.1118055, fu2025minority}.
Indeed, prior work has shown that human preferences do not always align with response correctness or objective quality~\cite{li-etal-2024-dissecting, sharma2024towards}.
These findings suggest that, as generated responses become more fluent and sophisticated, LLMs need to look beyond correctness and account for users' preferences and intentions~\cite{lin-etal-2024-interpretable, liu-etal-2025-user}.
Recent studies have therefore estimated user emotions or satisfaction from observable textual signals, such as user utterances, and used them to evaluate system responses~\cite{ghazarian-etal-2022-wrong}.
These approaches implicitly assume that users' affective states are reflected, at least to some extent, in their utterances and can be detected by third-party annotators or models.
However, even when users experience specific emotions, these internal affective states are not always explicitly expressed in text~\cite{katada-etal-2024-hazumi,koga-etal-2024-forecasting-implicit}.
Moreover, user satisfaction is not simply positive or negative sentiment, as it depends on whether users' needs, goals, or preferences are fulfilled~\cite{10.1145/3485447.3512020}.
Precisely because response generation quality has substantially improved, it is now important to understand satisfaction- or dissatisfaction-related user states from conversational interactions.

In this study, we investigate to what extent user dissatisfaction can be detected from subsequent user utterances in text-based chat interactions.
We treat dissatisfaction as an affective reaction that arises when a system response is perceived as failing to meet the user's intent, preferences, or needs.
As illustrated in Figure~\ref{fig:overview}, we formulate this as an utterance-level detection task using real ChatGPT dialogues.
We test multiple input conditions based on user utterances, including conditions that incorporate preceding system response or user utterance.
Using binary classification and anomaly detection, we further evaluate dissatisfaction detectability and explore how user dissatisfaction tends to appear in text.
The results indicate that user dissatisfaction can be detected to some extent, and suggest that behavioral and stylistic deviations from a prior state may provide informative cues for detecting dissatisfaction, particularly in anomaly detection.

\section{Related Work}
\subsection{System Evaluation in Interaction with AI}

Prior studies on human--AI interaction have widely adopted automatic evaluation methods that focus on system responses.
Such methods include automatic evaluators trained on human judgments~\cite{lowe-etal-2017-towards} and LLM-as-a-judge approaches~\cite{10.5555/3666122.3668142, lin-chen-2023-llm}.
In addition to evaluating positive aspects such as response quality, recent work has highlighted the importance of incorporating negative aspects such as dissatisfaction~\cite{abolghasemi2024cause}, and datasets annotated with users' own dissatisfaction have also been introduced~\cite{10.1145/3640543.3645148}. 
Several recent studies have further explored user utterances as signals of user reactions, because they often contain requests and emotions directed toward system responses~\cite{ghazarian-etal-2022-wrong, liu-etal-2025-user}.
Building on this line of research, this study focuses on dissatisfaction latent in user utterances.

\begin{figure*}
    \centering
    \includegraphics[width=0.83\linewidth]{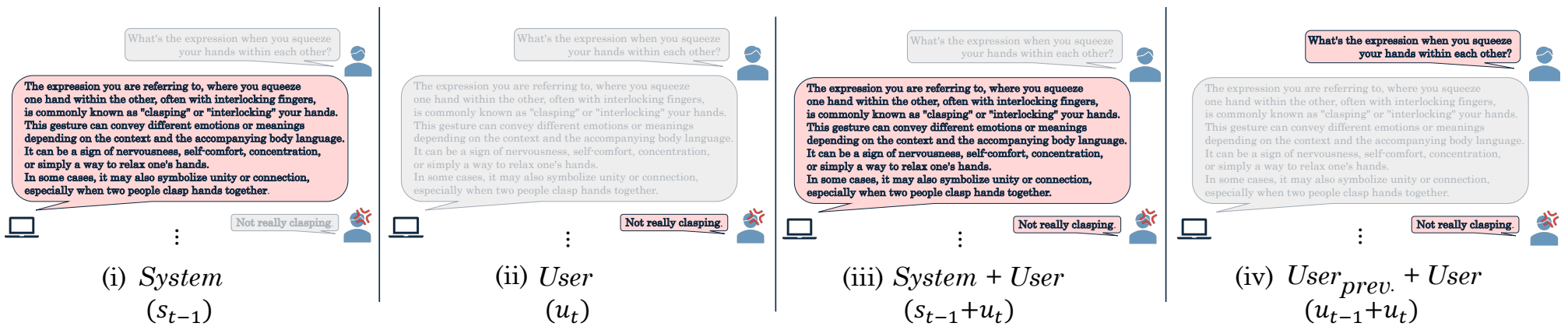}
    \caption{The four conditions used in the experiment. By examining these input conditions, we evaluate the effectiveness of detecting dissatisfaction using user utterances.}
    \vspace{-1em}
    \label{fig:input_condition}
\end{figure*}

\subsection{Emotion Estimation}

To improve dialogue quality, extensive research has been conducted on estimating emotions at the utterance level.
Representative methods include DialogueRNN~\cite{Majumder2019DialogueRNN},
fine-tuned models such as BERT and RoBERTa~\cite{Qin2023BERTERC}, and more recent LLM-based methods~\cite{feng-etal-2024-affect, hernandez-caralt-etal-2025-stupid}. 
Many of these studies rely on emotion labels provided by third-party annotators rather than by the users themselves~\cite{Busso2008IEMOCAP, poria-etal-2019-meld, li-etal-2017-dailydialog, sun2021simulating, see-manning-2021-understanding}. 
On the other hand, it has been reported that emotions perceived by third-party annotators often do not match users' actual emotions~\cite{truong2012speech, komatani-etal-2023-analyzing, numaya2025stylistic,kajiwara2021wrime}.
Since this study focuses on users' reactions to system responses rather than the correctness of the responses, we examine dissatisfaction detection within the framework of emotion estimation using self-reported labels.

\subsection{Emotional Expression in Human--AI Interaction}
Prior research in psychology suggests that communication with non-human agents may exhibit characteristics distinct from human--human interaction~\cite{gero2023social, mysore2025prototypical}, including the facilitation of negative emotion externalization~\cite{ho2018psychological}.
This property may provide clearer and more interpretable cues for analyzing negative effects, such as dissatisfaction, arising during dialogue with AI.

\section{Experimental Setup}
\label{sec:setup}
\subsection{Dissatisfaction Detection}

In this study, dissatisfaction is treated as an affective reaction that arises when a system response is perceived as failing to sufficiently meet the user's intent or preferences.
We investigate the extent to which such dissatisfaction can be detected from subsequent user utterances in text-based chat interactions.
Specifically, we used user utterances as input and treated labels derived from users’ self-reported dissatisfaction with the immediately preceding system responses as the ground truth.

\subsection{Dataset}

To analyze dissatisfaction at the level of individual user utterances, we used the ChatGPT Dissatisfaction Dataset~\cite{10.1145/3640543.3645148}, specifically its English subset, containing the largest number of annotated instances.
The dataset comprises interactions in which users labeled system responses from ChatGPT as dissatisfying.
In this study, a user utterance was labeled as \textit{dissatisfied} when it immediately followed a system response that the user had self-reported as dissatisfying, because such utterances could potentially reflect latent dissatisfaction; otherwise, it was labeled as \textit{non-dissatisfied}.
To ensure reliability, we conducted a preprocessing step to remove incomplete instances.
The final dataset contains 107 dialogues and 824 utterances, of which approximately 17.8\% are labeled as dissatisfied\footnote{Details of the dataset are provided in the Appendix \ref{appendix:dataset}.}.
Given the limited dataset size, five random splits were created, partitioning the data into training, validation, and test sets with a 6:1:3 ratio\footnote{Utterances from the same dialogue may appear in different splits; however, each instance uses only the current and at most one previous utterance. Therefore, this setup does not introduce direct contamination of extended dialogue context from the training data into the test data.}.

\subsection{Input Settings}
Let $u_t$ denote the user utterance at turn $t$ in a dialogue of length $L$ (total turns), and $s_t$ denote the corresponding system response.
The dialogue is represented as a sequence of utterance pairs $\{(u_t, s_t)\}_{t=1}^{L}$.
For dissatisfaction detection centered on $u_t$, four input conditions were defined, as illustrated in Figure~\ref{fig:input_condition}.
The \textit{System} uses only the immediately preceding system response $s_{t-1}$, for the comparison with conventional response-based methods~\cite{lowe-etal-2017-towards, 10.5555/3666122.3668142, lin-chen-2023-llm}.
The \textit{User} uses only $u_t$, to examine how much dissatisfaction is inferred from the user utterances.
The \textit{System+User} incorporates both $s_{t-1}$ and $u_t$ to interpret $u_t$ as a reaction to the preceding system response~\cite{ghazarian-etal-2022-wrong, liu-etal-2025-user}.
The \textit{User\textsubscript{prev.}+User} uses the pair of $u_{t-1}$ and $u_t$ to investigate whether dissatisfaction manifests as the change in user utterances.

\begin{table*}[t]
\centering
\renewcommand{\arraystretch}{0.88}
\small
\begin{tabular}{llcccc}
\toprule
& & \multicolumn{4}{c}{Input setting} \\
\cmidrule(lr){3-6}
Category & Method 
& \textit{System} 
& \textit{User} 
& \textit{System+User}  
& \textit{User\textsubscript{prev.}+User} \\
& 
& $(s_{t-1})$  
& $(u_t)$ 
& $(s_{t-1}+u_t)$ 
& $(u_{t-1}+u_t)$ \\
\midrule

\multirow{4}{*}{\textbf{Binary Classification}}
& BERT-small
& 32.0 $\pm$ 6.9
& 33.1 $\pm$ 2.7
& 33.0 $\pm$ 6.3 \inlinem{$-$0.1}
& 33.2 $\pm$ 3.1 \inlinep{$+$0.1} \\
& BERT-mini
& 31.1 $\pm$ 6.2
& 30.9 $\pm$ 2.9 
& 30.3 $\pm$ 7.3 \inlinem{$-$0.6}
& 26.4 $\pm$ 3.2 \inlinem{$-$4.5} \\
& GPT-5.1 (zero-shot)
& 20.3 $\pm$ 2.4
& 25.9 $\pm$ 2.5 
& 19.9 $\pm$ 0.7 \inlinem{$-$6.0}
& 23.7 $\pm$ 2.0 \inlinem{$-$2.2}\\
& GPT-5.1 (two-shot)
& 18.9 $\pm$ 1.1
& 26.1 $\pm$ 2.8 
& 20.7 $\pm$ 0.9 \inlinem{$-$5.4}
& 26.0 $\pm$ 3.1 \inlinem{$-$0.1} \\
\midrule

\multirow{8}{*}{\textbf{Anomaly Detection}}
& LOF (BERT)
& 24.3 $\pm$ 2.0
& 21.2 $\pm$ 4.2 
& 25.5 $\pm$ 4.2 \inlinep{$+$4.3}
& 22.0 $\pm$ 3.4 \inlinep{$+$0.8} \\
& LOF (GPT)
& 26.2 $\pm$ 1.2
& 20.0 $\pm$ 2.1 
& 27.5 $\pm$ 2.5 \inlinep{$+$7.5}
& 32.8 $\pm$ 4.6 \inlinep{$+$12.8} \\

& LUNAR (BERT)
& 21.4 $\pm$ 2.0
& 21.5 $\pm$ 2.9 
& 22.6 $\pm$ 2.8 \inlinep{$+$1.1}
& 24.0 $\pm$ 3.4 \inlinep{$+$2.5} \\
& LUNAR (GPT)
& 25.3 $\pm$ 3.6
& 20.1 $\pm$ 3.4 
& 28.6 $\pm$ 3.7 \inlinep{$+$8.5}
& 29.6 $\pm$ 7.8 \inlinep{$+$9.5} \\

& AE (BERT)
& 21.1 $\pm$ 1.6
& 22.7 $\pm$ 5.0 
& 22.7 $\pm$ 1.2 \inlinez{$\pm$0.0}
& 23.4 $\pm$ 2.1 \inlinep{$+$0.7} \\
& AE (GPT)
& 25.9 $\pm$ 2.3
& 20.5 $\pm$ 1.7 
& 25.9 $\pm$ 2.1 \inlinep{$+$5.4}
& 29.1 $\pm$ 3.6 \inlinep{$+$8.6} \\

& VAE (BERT)
& 20.4 $\pm$ 2.1
& 20.2 $\pm$ 4.1 
& 23.3 $\pm$ 1.8 \inlinep{$+$3.1}
& 21.2 $\pm$ 2.2 \inlinep{$+$1.0} \\
& VAE (GPT)
& 21.5 $\pm$ 1.1
& 19.1 $\pm$ 2.0 
& 22.0 $\pm$ 2.9 \inlinep{$+$2.9}
& 23.3 $\pm$ 2.4 \inlinep{$+$4.2} \\

\bottomrule
\end{tabular}
\caption{AUPRC for dissatisfaction detection (mean $\pm$ standard deviation over five random splits, \%). Values in parentheses indicate differences from the \textit{User} condition. 
For anomaly detection, the embedding model is shown in parentheses. 
The random baseline is 17.8\%, equal to the prevalence of dissatisfied utterances in the test data.}
\vspace{-0.5em}
\label{tab:all_methods_auprc}
\end{table*}

\subsection{Methods and Output}
To evaluate the extent to which dissatisfaction can be detected from user utterances, two approaches were employed: binary classification, commonly used for emotion estimation~\cite{mahato2024exploring, kubo2025developing}, and anomaly detection, which is suited to imbalanced data~\cite{barrett-etal-2022-lightweight}.

\paragraph{Binary Classification.}
In the binary classification setting, each $u_t$ was classified as either dissatisfied or non-dissatisfied.
Both BERT-based and GPT-based models, widely used in prior work, were adopted.
For BERT-based models, given the limited dataset size, 
which may be insufficient for fine-tuning large models~\cite{turc2019wellreadstudentslearnbetter, zhang2024scalingmeets, griesshaber2020finetuning}, compact variants—BERT-small\footnote{\url{https://huggingface.co/prajjwal1/bert-small}} and BERT-mini\footnote{\url{https://huggingface.co/prajjwal1/bert-mini}}—were fine-tuned~\cite{bhargava2021generalizationnliwaysnot, wasserblat2020exploring}.
These models output the probability of the dissatisfied class on the test set.
For GPT-based models, GPT-5.1 was used with prompting-based approaches, specifically zero-shot and two-shot settings, to classify utterances without additional training~\cite{mahato2024exploring, kubo2025developing}\footnote{Model configurations are provided in the Appendix \ref{appendix:model}.}.

\paragraph{Anomaly Detection.}
Given the class imbalance, with a relatively small proportion of dissatisfied utterances, anomaly detection was additionally performed by treating dissatisfied utterances as anomalies.
Specifically, inputs were converted into embeddings, and anomaly scores were computed in the resulting vector space. 
For embedding models, \texttt{bert-base-uncased}\footnote{\url{https://huggingface.co/google-bert/bert-base-uncased}} (BERT) and \texttt{text-embedding-3-large}\footnote{\url{https://developers.openai.com/api/docs/models/text-embedding-3-large}} (GPT) were used, following recent findings that embedding-based anomaly detection methods are effective in NLP and that both BERT-based and OpenAI embeddings achieve competitive performance~\cite{li-etal-2025-nlp}.
We used Local Outlier Factor (LOF)~\cite{10.1145/342009.335388}, LUNAR (Unifying Local Outlier Detection Methods via Graph Neural Networks)~\cite{goodge2022lunar}, Autoencoder (AE)~\cite{aggarwal2016introduction}, and Variational Autoencoder (VAE)~\cite{kingma2013auto, burgess2018understanding}\footnote{Details of the methods are provided in the Appendix \ref{appendix:anomaly detection method}.}.
During evaluation, anomaly scores were assigned to the test set.

\subsection{Evaluation Metric}
Given the class imbalance, the area under the precision--recall curve (AUPRC) was adopted.
For AUPRC, the positive class prevalence corresponds to the expected performance of a random baseline~\cite{saito2015precision}.
Because dissatisfied utterances account for 17.8\% of the dataset used in this study, 17.8\% was used as the reference value for interpreting AUPRC.

\section{Results and Analysis}
\label{main result}
Using the methods described in Section~\ref{sec:setup}, we examined whether user dissatisfaction can be detected from user utterances and analyzed which input conditions facilitate such detection.
Each method was evaluated on all five random stratified splits.
Table~\ref{tab:all_methods_auprc} reports the mean and standard deviation of the AUPRC across the five splits.

\subsection{Dissatisfaction Detection Performance}
\label{sec:user_utter}
The results in the \textit{User} column of Table~\ref{tab:all_methods_auprc} show the score of dissatisfaction detection when only $u_t$ was used as input.
Across all binary classification and anomaly detection methods, AUPRC consistently exceeded the dissatisfaction prevalence.
These findings indicate that user dissatisfaction can be detected to some extent.
However, overall performance remained moderate, suggesting that user utterances do not consistently reflect users' latent affective states in a detectable form.

The \textit{System} column in Table~\ref{tab:all_methods_auprc} reports scores when only $s_{t-1}$ was used as input.
In the binary classification setting, \textit{User} outperformed \textit{System} across all models except BERT-mini, and the difference was statistically significant for the GPT-based models\footnote{Full significance test results are provided in Appendix~\ref{app:significance_tests}.}.
This suggests that directly modeling user reactions can be generally more useful for dissatisfaction detection than relying solely on system responses.
In contrast, in anomaly detection, \textit{User} yielded lower scores than \textit{System} for many methods, indicating that the effectiveness of \textit{User} inputs depends on the modeling approach.

\subsection{Effect of Adding System Response}
\label{sec:add_system}
The \textit{System+User} column in Table~\ref{tab:all_methods_auprc} reports scores when $s_{t-1}$ was added to $u_t$ as input.
The \textit{System+User} condition was often comparable to or better than the \textit{System} condition, perticularly in anomaly detection.
This suggests that information contained in user utterances contributes to effective dissatisfaction detection.
In the binary classification setting, AUPRC scores were lower than those of the \textit{User} condition, which used only $u_t$, with larger and statistically significant declines observed for the GPT-based models.
To examine this effect, Table~\ref{tab:qual_example_cond2_A} presents an example in which only the \textit{System+User} condition incorrectly classified a non-dissatisfied utterance as dissatisfied in GPT-based models.
In this example, $u_t$ lacks contextual coherence with $s_{t-1}$, and the resulting topic shift may be misinterpreted as a signal of dissatisfaction.
This interpretation is consistent with prior findings that dissatisfied users shift topics or change request types during interactions with LLMs~\cite{10.1145/3640543.3645148}.
However, as this example illustrates, natural topic shifts can also occur without dissatisfaction.
These findings suggest that while incorporating preceding context can provide useful cues for dissatisfaction detection, it may also introduce errors for certain utterances.

\begin{table}[t]
\centering
\small
\renewcommand{\arraystretch}{0.93}
\begin{tabularx}{\linewidth}{l Y}
\toprule
 \multicolumn{2}{l}{Interaction} \\
\midrule
$s_{t-1}$ & \textit{To implement the search() method on the client side, you'll need to communicate with the server ... } \\
\rowcolor{gray!12}
$u_t$     & \textit{displaySearchResults(results)} \\
\bottomrule
\end{tabularx}
\caption{Example of a non-dissatisfied utterance that is incorrectly classified as dissatisfied under \textit{System+User} in binary classification setting.}
\vspace{-1em}
\label{tab:qual_example_cond2_A}
\end{table}

\subsection{Effect of the Previous User Utterance}
\label{sec:add_user}
The \textit{User\textsubscript{prev.}+User} column in Table~\ref{tab:all_methods_auprc} reports scores when $u_{t-1}$ was added to $u_t$.
In the binary classification setting, AUPRC scores were lower than those of the \textit{User}, but remained higher than the \textit{System} condition, with statistically significant differences observed for the GPT-based models.
In anomaly detection, \textit{User\textsubscript{prev.}+User} achieved the highest scores among almost all input settings, with statistically significant differences observed for several comparisons. 
At the same time, incorporating the preceding user utterance did not improve performance in the binary classification setting.
These results suggest that, particularly for anomaly detection, certain signals of dissatisfaction are difficult to detect from $u_{t}$ alone but become more detectable when changes relative to $u_{t-1}$ are considered.
Based on this finding, Section~\ref{style_analysis} focuses on dissatisfied utterances that are more readily detected under the \textit{User\textsubscript{prev.}+User} condition and analyzes their characteristics.

\section{Discussion: Why Previous User Utterances Help Detect Dissatisfaction}
\label{style_analysis}

\subsection{Comparison Setup}
\paragraph{Hypotheses.}

Based on the results in Section~\ref{main result}, anomaly detection with GPT-based embeddings shows larger improvement when using \textit{User\textsubscript{prev.}+User} compared with \textit{User}.
Two possible explanations are considered: (i) even when dissatisfaction is expressed in $u_t$, the information of $u_t$ alone may be insufficient to fully capture its context; and (ii) dissatisfaction may become more detectable when changes in user behavior between $u_t$ and $u_{t-1}$ are taken into account.
To examine these hypotheses, utterance length is analyzed as a proxy for the amount of information contained in each utterance (Section~\ref{utterance_length}), and changes in style vectors are analyzed to capture behavioral shifts (Section~\ref{style_vector}).

\begin{table}[t]
\centering
\small
\renewcommand{\arraystretch}{0.93}
\begin{tabular}{lcc}
\toprule
Metric
& \shortstack[c]{\textit{Context-}\\\textit{advantaged}}
& \shortstack[c]{\textit{Consistently}\\\textit{high-scoring}}
\\
\midrule

Utterance length ($u_t$)
& 10.4
& 27.6
\\
\midrule

Style vector \\
\quad Cosine similarity
& 0.741
& 0.932
\\
\quad $\Delta$ $L_2$
& 10.56
& 5.84
\\
\bottomrule
\end{tabular}
\caption{Comparison of utterance length and style features. High-scored in \textit{$u_{t-1} + u_t$} cases (left) vs. consistently detectable dissatisfaction utterances (right).}
\label{tab:analysis_length_style}
\end{table}

\paragraph{Subsets for Hypothesis Testing.}
To test these hypotheses, we picked up two subsets of dissatisfied utterances under the following conditions and compared them to identify patterns for which incorporating the previous user utterance is particularly effective in detecting dissatisfaction.
The first set, \textit{context-advantaged}, consists of dissatisfied utterances that fall within the top 20\% of anomaly scores under \textit{User\textsubscript{prev.}+User} and whose anomaly scores exceed those under \textit{User}.
The second set, \textit{consistently high-scoring}, consists of dissatisfied utterances that fall within the top 20\% of anomaly scores across \textit{User}, \textit{System+User}, and \textit{User\textsubscript{prev.}+User}.
The first set contains 30 dissatisfied utterances, and the second set contains 24 dissatisfied utterances.

\subsection{Utterance Length}
\label{utterance_length}

To examine Hypothesis (i), differences in the word length of $u_t$ were analyzed.
As shown in Table~\ref{tab:analysis_length_style} (\textit{utterance length}), the average length of $u_t$ in the \textit{context-advantaged} set was 10.4 words, which was shorter than the average of 27.6 words in the \textit{consistently high-scoring} set.
This difference was statistically significant according to the Mann-Whitney U test~\cite{497e1044-d5b0-30a9-b230-3ca0f10d6f6c,aldihan2022pilot, dawkins2025detection} ($p < 0.05$).
Table~\ref{tab:qual_example_cond3_style} presents an example in which dissatisfaction becomes easier to detect under the \textit{User\textsubscript{prev.}+User} condition, particularly when $u_t$ is short.
In this case, $u_t$ alone conveyed limited information; however, incorporating $u_{t-1}$ provided contextual cues that facilitate interpretation of $u_t$ as expressing dissatisfaction with the preceding response or as an additional request.
These findings suggest that, for dissatisfied utterances correctly identified as anomalies when $u_{t-1}$ is included, the standalone utterance $u_t$ may lack sufficient cues, and the inclusion of $u_{t-1}$ compensates by providing contextual information.

\begin{table}[t]
\centering
\small
\renewcommand{\arraystretch}{0.93}
\begin{tabularx}{\linewidth}{l Y}
\toprule
\multicolumn{2}{l}{Interaction} \\
\midrule
\rowcolor{gray!12}
$u_{t-1}$ & \textit{\textbf{INSTAGRAM BIO FOR A APPLE WATCH SHOP'S INSTGRAM}} \\

$s_{t-1}$ & \textit{Timekeepers of Innovation Your destination for all things Apple Watch! Discover the perfect ...} \\

\rowcolor{gray!12}
$u_t$     & \textit{\textbf{SHORTER}} \\
\bottomrule
\end{tabularx}
\caption{Example of a dissatisfied utterance with a short $u_t$, which improves detection under \textit{User\textsubscript{prev.}+User} compared to other conditions.}
\label{tab:qual_example_cond3_style}
\end{table}

\subsection{Stylistic Changes}
\label{style_vector}
To examine Hypothesis (ii), changes in the user style between $u_{t-1}$ and $u_t$ were analyzed using word style vectors~\cite{akama2018stylevec}, which represent stylistic properties in an embedding space.
Specifically, an utterance-level style vector was computed by averaging the word style vectors within each utterance.
Style changes were then quantified using cosine similarity and the L2 norm of the difference vector between $u_{t-1}$ and $u_t$.

As shown in Table~\ref{tab:analysis_length_style} (\textit{style vector}), the \textit{context-advantaged} set exhibited lower cosine similarity and larger $\Delta L_2$ than the \textit{consistently high-scoring} set.
These differences were statistically significant according to the Mann-Whitney U test (cosine similarity: $p < 0.05$, $\Delta L_2$: $p < 0.05$).
These results indicate that the \textit{context-advantaged} set is characterized by greater stylistic shifts between consecutive user utterances.
This suggests that capturing such behavioral changes contributes to dissatisfaction detection, and that anomaly detection methods may be particularly well-suited to capturing these signals.

\begin{table}[t]
\centering
\small
\renewcommand{\arraystretch}{0.93}
\begin{tabularx}{\linewidth}{l l Y}
\toprule
\textbf{Interaction} & \cellcolor{gray!12}${u_{t-1}}$ & \cellcolor{gray!12}\textit{How do i uninstall wgcebas from my pc} \\
& ${s_{t-1}}$ & \textit{I'm not familiar with a program called "wgcebas," so I can't provide specific instructions for uninstalling it. However, I can give you general guidance on how to uninstall software from your PC...} \\
& \cellcolor{gray!12}${u_{t}}$ &  \cellcolor{gray!12}\textit{Isnt there other way to delete it} \\
\midrule
\textbf{Scores} & \multicolumn{2}{l}{Cosine similarity: 0.49, $\Delta L_2$: 15.59} \\
\bottomrule
\end{tabularx}
\caption{Example of a dissatisfied utterance with observable stylistic changes between $u_{t-1}$ and $u_t$, which improves detection under \textit{User\textsubscript{prev.}+User} compared to other conditions.}
\label{tab:qual_example_cond3_style_change}
\end{table}

Table~\ref{tab:qual_example_cond3_style_change} presents an example in which dissatisfaction becomes easier to detect under the \textit{User\textsubscript{prev.}+User} condition, accompanied by a change in user style.
In this case, $u_t$ exhibited stylistic shifts, such as lexical omission, relative to $u_{t-1}$, and the cosine similarity (0.49) was lower than the corresponding values reported in Table~\ref{tab:analysis_length_style}.
This suggests that modeling behavioral changes between $u_{t-1}$ and $u_t$ facilitates interpretation of the utterance as expressing dissatisfaction with the preceding system response.

\begin{table}[t]
\centering
\small
\renewcommand{\arraystretch}{0.93}
\begin{tabularx}{\linewidth}{l l Y}
\toprule
\textbf{Interaction} & \cellcolor{gray!12}${u_{t-1}}$ & \cellcolor{gray!12}\textit{10 more} \\
& ${s_{t-1}}$ & \textit{Certainly! Here are ten more unusual uses for a brick:1. DIY stepping stones: Use bricks as molds for creating custom stepping stones...} \\
& \cellcolor{gray!12}${u_{t}}$ &  \cellcolor{gray!12}\textit{5 more} \\
\midrule
\textbf{Scores} & \multicolumn{2}{l}{Cosine similarity: 0.96, $\Delta L_2$: 2.60} \\
\bottomrule
\end{tabularx}
\caption{Example of a dissatisfied utterance that receives a low anomaly score in anomaly detection but is correctly detected by binary classification, where dissatisfaction is expressed as repetition of similar requests.}
\label{tab:qual_example_cond3_repeat}
\end{table}

\subsection{Challenges of Stylistic Change}
\label{sec:differernce_methods}
As discussed above, stylistic changes between consecutive user utterances can contribute to dissatisfaction detection, particularly in anomaly detection.
In contrast, dissatisfaction may also appear without clear stylistic changes, making it difficult to detect using anomaly detection.
In this subsection, we analyze such cases under the \textit{User\textsubscript{prev.}+User} condition by comparing with binary classification.

Table~\ref{tab:qual_example_cond3_repeat} presents an example in which dissatisfaction was difficult to detect using anomaly detection, but was correctly identified by binary classification.
In this case, similar requests are repeated in $u_{t-1}$ and $u_t$, and the cosine similarity was high (0.96), indicating highly similar utterance styles.
Prior work has shown that users can signal dissatisfaction by repeatedly rephrasing the same query~\cite{lin-etal-2024-interpretable, 10.1145/3640543.3645148}.
In this example, dissatisfaction is expressed through repetition rather than stylistic change, making it difficult to detect by anomaly detection.
Consequently, anomaly detection methods that rely on deviations from the previous utterance, as discussed in Section~\ref{style_vector}, fail to capture dissatisfied utterances with minimal change.
These findings suggest that dissatisfaction can be expressed in multiple forms, including those that rely on cues beyond stylistic changes.

\section{Evaluation in Multilingual Settings}
\label{multilingual}

The expression of user dissatisfaction may vary across languages and cultural contexts. As an exploratory analysis, we therefore examine how the English-based findings in Section~\ref{main result} appear in three additional languages, Polish, Spanish and Japanese.

\begin{table}[t]
\centering
\small
\renewcommand{\arraystretch}{0.88}
\begin{tabular}{l r r r}
\toprule
Language & Dialogues & Utterances & \makecell{Dissatisfied \\ Utterances (\%)} \\
\midrule
English & 107 & 824 & 147 (17.8) \\
Polish & 20 & 152 & 20 (13.2) \\
Spanish & 6 & 29 & 6 (20.7) \\
Japanese & 26 & 192 & 50 (26.1) \\
\bottomrule
\end{tabular}
\caption{Statistics of the multilingual datasets.}
\label{tab:dataset_multilingual}
\end{table}

\paragraph{Experimental Setup.}

In the ChatGPT Dissatisfaction Dataset, we selected Polish and Spanish as additional non-English languages, as they were among the largest available subsets after English.
In addition, to increase linguistic diversity in the multilingual setting beyond languages written primarily in the Latin alphabet, we newly collected Japanese dialogue data\footnote{\url{https://crowdsourcing.yahoo.co.jp/}} following the original data collection procedure and annotated them with self-reported dissatisfaction labels.
All languages were preprocessed using the same steps as the English data. 
Dataset statistics are shown in Table~\ref{tab:dataset_multilingual}.
As the datasets for each language are smaller than the English dataset and insufficient for training, anomaly detection using LOF was applied only at test time to examine the expression of dissatisfaction.
For BERT-based embeddings, we used language-specific pretrained BERT models for Polish, Spanish, and Japanese\footnote{\raggedright Polbert: \url{https://huggingface.co/dkleczek/bert-base-polish-uncased-v1}\\
BETO: \url{https://huggingface.co/dccuchile/bert-base-spanish-wwm-uncased}\\
BERT base Japanese: \url{https://huggingface.co/tohoku-nlp/bert-base-japanese-v3}}, while we used \texttt{text-embedding-3-large} for GPT-based embeddings, following the English experiments\footnote{Details of the settings are provided in Appendix~\ref{appendix:anomaly detection multilingual}.}.

\begin{table}[t]
\centering
\small
\renewcommand{\arraystretch}{0.93}
\setlength{\tabcolsep}{3pt} %
\begin{tabular}{lcccc}
\toprule
& \multicolumn{4}{c}{Input setting} \\
\cmidrule(lr){2-5}
& \textit{System} 
& \textit{User} 
& \makecell[c]{\textit{System}\textit{+User}}  
& \makecell[c]{\textit{User\textsubscript{prev.}}\textit{+User}} \\
\midrule

\multicolumn{5}{l}{\textbf{Polish}}  \\
\quad BERT
& 31.4
& 16.8 
& 30.0 \inlinep{$+$13.2}
& 15.9 \inlinem{$-$0.9} \\
\quad GPT
& 32.8
& 15.9
& 37.5 \inlinep{$+$21.6}
& 16.7 \inlinep{$+$0.8} \\
\midrule

\multicolumn{5}{l}{\textbf{Spanish}}  \\
\quad BERT
& 55.2
& 38.5 
& 44.2 \inlinep{$+$5.7}
& 23.5 \inlinem{$-$15.0} \\
\quad GPT
& 34.5
& 25.6
& 41.0 \inlinep{$+$15.4}
& 30.6 \inlinep{$+$5.0} \\
\midrule

\multicolumn{5}{l}{\textbf{Japanese}} \\
\quad BERT
& 35.9
& 30.5 
& 35.8 \inlinep{$+$5.3}
& 26.2 \inlinem{$-$4.3} \\
\quad GPT
& 33.3
& 31.7 
& 35.4 \inlinep{$+$3.7}
& 30.7 \inlinem{$-$1.0} \\
\bottomrule
\end{tabular}
\caption{AUPRC for multilingual dissatisfaction detection using LOF under each condition and embedding model (\%). Values in parentheses indicate differences from the \textit{User} condition. The random baseline is 13.2\% for Polish, 20.7\% for Spanish, and 26.1\% for Japanese.}
\label{tab:multilingual_lof_auprc}
\end{table}

\paragraph{Results and Discussion.}
Table~\ref{tab:multilingual_lof_auprc} reports the performance of dissatisfaction detection on multilingual datasets.
The advantage of the \textit{User\textsubscript{prev.}+User} condition observed in the English experiments was not evident in the multilingual setting.
In contrast, the \textit{User} condition yielded relatively lower performance in the multilingual settings, and the \textit{System+User} conditions achieved higher scores.
Additionally, Polish and Spanish showed the similar tendency across conditions.
These results suggest that the expression of dissatisfaction may vary across languages and cultures.
While changes relative to $u_{t-1}$ provided useful cues in English, such cues may be less accessible from user utterances in these languages.
These findings underscore the need for careful analysis when extending dissatisfaction detection to other languages.
Future work should begin by developing larger multilingual datasets.
This would make it possible to apply the same training-based anomaly detection used in the English experiments to other languages, allowing comparisons under more consistent methodological conditions.
It would also help distinguish the effects of dataset size and methodology from language- and culture-related differences in dissatisfaction detection.

\begin{table}[t]
\centering
\small
\renewcommand{\arraystretch}{0.9}
\begin{tabularx}{\linewidth}{l Y}
\toprule
\multicolumn{2}{l}{Interaction} \\
\midrule
\rowcolor{gray!12}
$u_{t-1}$     & \textit{oblicz liczbe dostanych mandatow uzywajac tych danych: Partia A = 720, Partia B = 300, Paria C = 480 uzywajac metody d'hondta} \\
\rowcolor{gray!12}
& {\footnotesize (Calculate the number of seats received using these data: Party A = 720, Party B = 300, Party C = 480 using the d'Hondt method)} \\
\midrule
$s_{t-1}$ & \textit{D'Hondta jest metodą wyborczą, która służy do podziału mandatów między partiami lub kandydatami w wyborach. Aby obliczyć \textbf{liczbę mandatów} ...} \\
& {\footnotesize (The D'Hondt method is an electoral method used to allocate seats between parties or candidates in elections. To calculate the number of seats...)} \\
\midrule
\rowcolor{gray!12}
$u_t$     & \textit{oblicz \textbf{liczbe} dostanych \textbf{mandatow} uzywajac tych danych: Partia A = 720, Partia B = 300, Partia C = 480} \\
\rowcolor{gray!12}
& {\footnotesize (Calculate the number of seats received using these data: Party A = 720, Party B = 300, Party C = 480)} \\
\bottomrule
\end{tabularx}
\caption{Example of a dissatisfied utterance with similar content to the preceding utterance in the Polish dataset. English translations are provided in parentheses.}
\vspace{-1em}
\label{tab:qual_example_polish_repeat}
\end{table}

\paragraph{Qualitative Analysis.}

In anomaly detection, \textit{User\textsubscript{prev.}+User} achieved the best performance in English, but not in the multilingual conditions. 
For the qualitative analysis, we focus on Polish and Japanese, which constitute the two largest non-English datasets in this study.
To examine whether cues observed in English differ across languages, the analysis focused on repeated similar requests, which served as a cue for dissatisfaction in the English dataset, as discussed in Section~\ref{sec:differernce_methods}.
Examples with similar characteristics were analyzed in the Polish and Japanese datasets.

Table~\ref{tab:qual_example_polish_repeat} shows an example from the Polish data in which the dissatisfied utterance $u_t$ was not correctly detected.
In this case, the user repeats nearly the same request, suggesting dissatisfaction with the system response not matching the user's intent.
It also illustrates that diacritical marks present in the system response were omitted in the user utterance, as shown in the boldfaced parts of Table~\ref{tab:qual_example_polish_repeat}.
Such orthographic variation has been reported in user-generated utterances in many languages that use diacritics~\cite{adali2014vowel, ljubevsic2016corpus, krstev2018knowledge}, and may hinder dissatisfaction detection~\cite{gorman-pinter-2025-dont}.
Furthermore, another possible reason for the difficulty in detecting dissatisfaction is that Polish is known to be morphologically complex and relatively low-resource~\cite{wolk2021real, mroczkowski2021herbert}.

Table~\ref{tab:qual_example_japanese_repeat} shows an example from the Japanese data in which a similar interaction pattern does not necessarily indicate dissatisfaction depending on the user.
This example illustrates that user-utterance patterns that can function as cues to dissatisfaction in English do not necessarily correspond to dissatisfaction in the Japanese dataset.
One possible interpretation is that dissatisfaction may be expressed less explicitly in Japanese, given reports that Japanese speakers tend to show lower self-disclosure than English speakers~\cite{kito2005self}.
Overall, while the repetition of utterances serves as a cue for dissatisfaction, these findings suggest that the effectiveness of this feature may vary across linguistic and cultural contexts. They also leave open the possibility that individual-level factors may affect how such cues appear in user utterances.
More systematic analyses are therefore needed to examine these possibilities in future work.

\begin{table}[t]
\centering
\small
\renewcommand{\arraystretch}{0.9}
\begin{tabularx}{\linewidth}{l Y}
\toprule
 \multicolumn{2}{l}{Interaction} \\
\midrule
\rowcolor{gray!12}
$u_{t-1}$     & \textnormal{地上波だよ} \\
\rowcolor{gray!12}
& (Terrestrial television) \\
\midrule
$s_{t-1}$ & \textnormal{調べたけど、2025年時点では「美しい彼（うつくしい彼）」は地上波での定期再放送予定が確認されていない、という情報が出ています。 ... 全話無料視聴できるサブスクサービス一覧 ...} \\
& (I checked, and it seems that as of 2025, there are no confirmed plans for regular rebroadcasts of ``Utsukushii Kare'' (Beautiful Man) on terrestrial television. ... List of subscription services where you can watch all episodes for free ...) \\
\midrule
\rowcolor{gray!12}
$u_t$     & \textnormal{違うよ 現在地上波で放送中のやつだよ} \\
\rowcolor{gray!12}
& (No, that's the one currently airing on terrestrial television.) \\
\bottomrule
\end{tabularx}
\caption{Example of a non-dissatisfied utterance with similar content to the preceding utterance in the Japanese dataset. English translations are provided in parentheses.}
\vspace{-1em}
\label{tab:qual_example_japanese_repeat}
\end{table}

\section{Conclusion}
This study investigated whether user dissatisfaction in human--AI interactions can be detected from user utterances.
We formulated dissatisfaction detection as a task using self-reported dissatisfaction labels as ground truth, then examined it with both binary classification and anomaly detection.
The results indicate that user utterances contain informative cues for dissatisfaction detection, while also suggesting that such dissatisfaction is not always explicitly stated. 
Moreover, our analysis suggests that incorporating the immediately preceding user utterance can improve detection performance, particularly for anomaly detection in English. 
This finding indicates that changes relative to a user’s prior expression can provide useful signals beyond the current utterance alone.
Furthermore, experimental results in multilingual settings indicate that informative cues for dissatisfaction detection vary across languages, highlighting the importance of accounting for language-specific and culture-specific characteristics.

\clearpage

\section*{Limitations}

First, the scale of the data used for analysis is limited. 
This constraint partly reflects the relatively recent emergence of LLMs as collaborative agents in human-centered tasks, and the corresponding lack of well-developed research resources. 
Moreover, dialogue datasets in which speakers annotate their own utterances with emotions or dissatisfaction at the utterance level remain scarce.
Given the limited dataset size, we used utterance-level random splits rather than dialogue-level splits. 
Each instance uses only the current and at most one immediately preceding utterance, which limits direct leakage of extended dialogue context across splits. 
However, utterances from the same dialogue may still appear in different splits.
With a larger dataset, future work could adopt dialogue-level splits to provide a stricter evaluation of generalization to unseen dialogues.
This limitation is particularly pronounced in the multilingual analysis, where the non-English datasets were substantially smaller than the English dataset and were therefore used as a preliminary case study.
As a result, the present study does not fully disentangle whether the differences observed across languages reflect language- and culture-related differences or the effects of dataset size.

Second, the notion of ``dissatisfaction'' examined in this study is inherently subjective and dependent on individual users, making it difficult to define and operationalize in a uniform manner. 
Even when identical system responses are presented, users may differ in whether they perceive them as dissatisfying. 
Furthermore, there is substantial variation in the extent to which users explicitly verbalize their emotions in text. 
Consistent with this, the dataset used in this study includes instances where utterances with highly similar content are annotated as expressing dissatisfaction in some cases, while being judged as non-dissatisfaction in others. 
Note that dissatisfied utterances in this study are not restricted to cases in which dissatisfaction is explicitly expressed in the text.

Finally, this study focuses on open-ended human--LLM communication, primarily based on the ChatGPT Dissatisfaction Dataset. 
As such, the findings derived from this specific setting may not necessarily generalize directly to task-oriented dialogue systems or other conversational environments.

\section*{Ethical Considerations}
One experiment in this study involved human participants. 
To collect Japanese-language data for multilingual evaluation, native Japanese speakers were recruited via a crowdsourcing platform, Yahoo! Crowdsourcing.
Prior to participation, participants received detailed instructions describing the task structure and procedures. 
Informed consent was obtained for data collection, research use, and potential future public release of the dataset. Participants also agreed to share their dialogue data in accordance with the applicable terms of service governing their use of ChatGPT.
Task completion required approximately 10 minutes. 
Compensation was set to the legally mandated minimum wage in Tokyo (JPY 1,226 per hour), and each participant received JPY 205 for completing the task.

The collected data may include potentially sensitive content, such as opinions related to political preferences or social values; however, all data were used exclusively for dialogue evaluation analysis. 
Appropriate measures were taken to protect participant privacy and to ensure compliance with relevant ethical guidelines.

\section*{Acknowledgments}
This research was supported in part by JSPS KAKENHI Grant Number JP25K21263, JP26H02511, JST BOOST Grant Number JPMJBY24A1, JST Moonshot R\&D Program Grant Number JPMJMS2011-35 (fundamental research), and the ``R\&D Hub Aimed at Ensuring Transparency and Reliability of Generative AI Models'' project of the Ministry of Education, Culture, Sports, Science and Technology.

\bibliography{references}

\clearpage
\appendix

\section{Details of Dataset}
\label{appendix:dataset}

\subsection{Preprocess} 
To ensure the reliability of the analysis, the following preprocessing steps were applied to the data used in this study:
\begin{enumerate}
    \item \textbf{Removal of incomplete data.} Dialogues in which all utterances were masked, as well as instances with missing dialogue logs, were excluded from the dataset.
    \item \textbf{Removal of task-incompatible data.} Because the target of analysis is dissatisfaction with system responses, cases in which no user utterance followed a system response labeled as dissatisfying and the dialogue ended immediately after that response were excluded. Such cases accounted for 50 of 157 dialogues in English and 7 of 27 dialogues in Polish.
    \item \textbf{Securing contextual information.} Because some input settings required utterances preceding $u_t$, first-turn utterances without relevant prior context were excluded.
\end{enumerate}

\begin{table}[ht]

\centering
\footnotesize
\setlength{\tabcolsep}{6pt}
\renewcommand{\arraystretch}{0.93}

\begin{tabular}{p{0.13\linewidth} p{0.75\linewidth}}
\toprule

\textit{System} &
{I apologize for the confusion. \ldots} \\
\midrule
\textit{User}&
{It's longer than the original.} \\
\midrule
\textit{System + User} &
{\texttt{[ASSISTANT]} I apologize for the confusion. \ldots}\\
& {\texttt{[USER]} It's longer than the original.} \\
\midrule
\textit{User-Previous + User} &
{\texttt{[PREV]} That is even longer in terms of characters.}\\
& {\texttt{[CUR]} It's longer than the original.} \\

\bottomrule
\end{tabular}
\caption{Example of four conditions.}
\label{tab:input_data_example}
\end{table}

\subsection{Input Formatting and Examples}
For the BERT-based models, the input text was tokenized with a maximum length of \texttt{max\_length}.
In addition, to preserve information from the $u_t$ placed at the end of the input sequence, the truncation direction was set to the left (\texttt{truncation\_side=left}).
Experiments were conducted under four different input settings.
In \textit{System+User}, to make speaker boundaries explicit, \texttt{[ASSISTANT]} was inserted  before $s_{t-1}$ and \texttt{[USER]} before $u_t$.
In \textit{User\textsubscript{prev.}+User}, to capture inter-utterance relations using special tokens~\cite{wu-etal-2020-tod, 9678172, chen2021dsbertunsuperviseddialoguestructurelearning}, \texttt{[PREV]} was inserted before $u_{t-1}$ and \texttt{[CUR]} before $u_t$.
Examples are shown in Table~\ref{tab:input_data_example}.

\section{Details of Binary Classification}
\label{appendix:model}

\subsection{Model Selection}
\label{subsec:choose_model}

Scaling laws suggest that performance depends on both model size and data size, and that under data-constrained settings, larger models are not necessarily optimal~\cite{zhang2024scalingmeets}.
It has also been shown that compact BERT-based models, when fine-tuned after pretraining, can achieve performance comparable to that of larger models~\cite{turc2019wellreadstudentslearnbetter}.
Furthermore, fine-tuning is expected to be difficult when the number of training instances is fewer than 1,000~\cite{griesshaber2020finetuning}.

\begin{figure}[ht]
\centering
\includegraphics[width=0.95\linewidth]{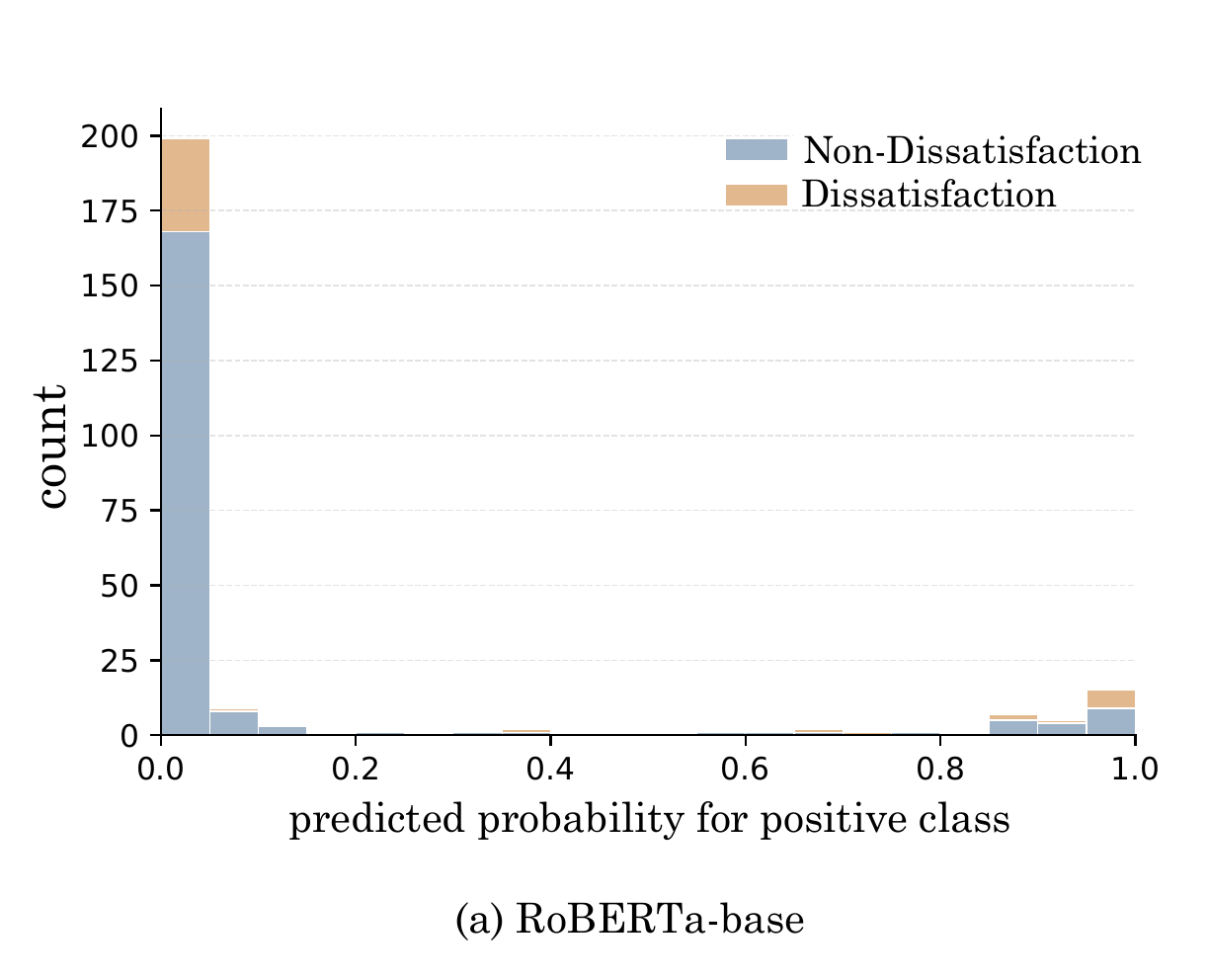}
\caption{Distribution of predicted probabilities for binary classification with RoBERTa-base.}
\label{fig:roberta}
\end{figure}

\begin{figure}[ht]
\centering
\includegraphics[width=0.95\linewidth]{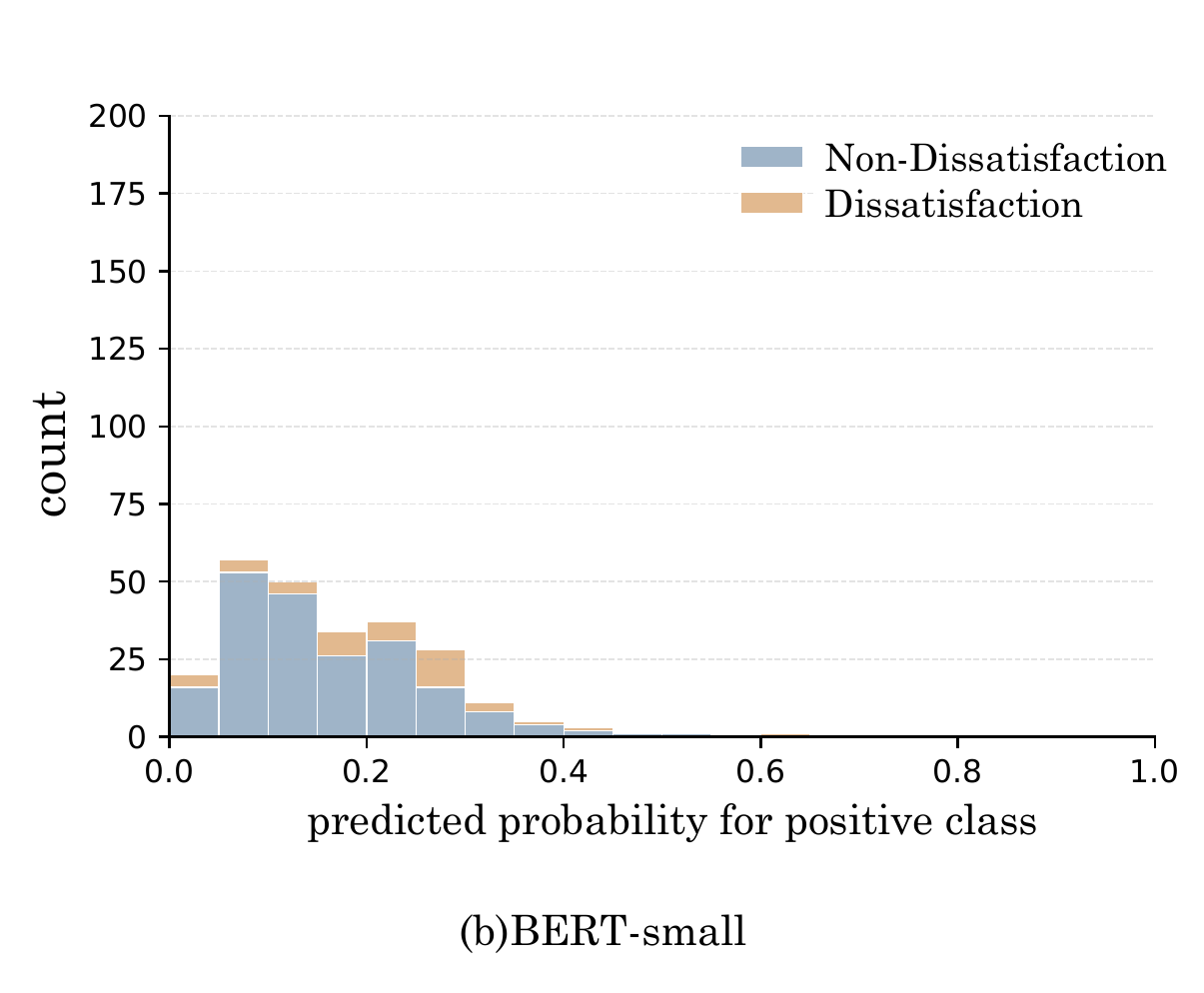}
\caption{Distribution of predicted probabilities for binary classification with BERT-small.}
\label{fig:bert_small}
\end{figure}

As shown in Table~\ref{tab:dataset_multilingual} in the main text, the dataset used in this study is relatively small.
We therefore examined whether training could be conducted appropriately on this dataset across multiple models with different numbers of parameters.
Here, Figures~\ref{fig:roberta} and~\ref{fig:bert_small} show the behavior of predicted probabilities on the validation data for the relatively large RoBERTa-base and the more lightweight BERT-small, respectively.
As shown in Figure~\ref{fig:roberta}, when binary classification was performed using RoBERTa-base, the predicted probabilities on the validation data were concentrated near 0 for almost all instances, indicating that the model failed to learn effectively.
In contrast, as shown in Figure~\ref{fig:bert_small}, the predicted probabilities produced by BERT-small were more dispersed.
Therefore, in this study, BERT-small and BERT-mini were used for dissatisfaction expression detection, taking the dataset size into account.

\subsection{Prompts Used for GPT-based Models}
\label{subsec:appendix_gpt_model}

\begin{figure}[ht]
\centering
\includegraphics[width=0.95\linewidth]{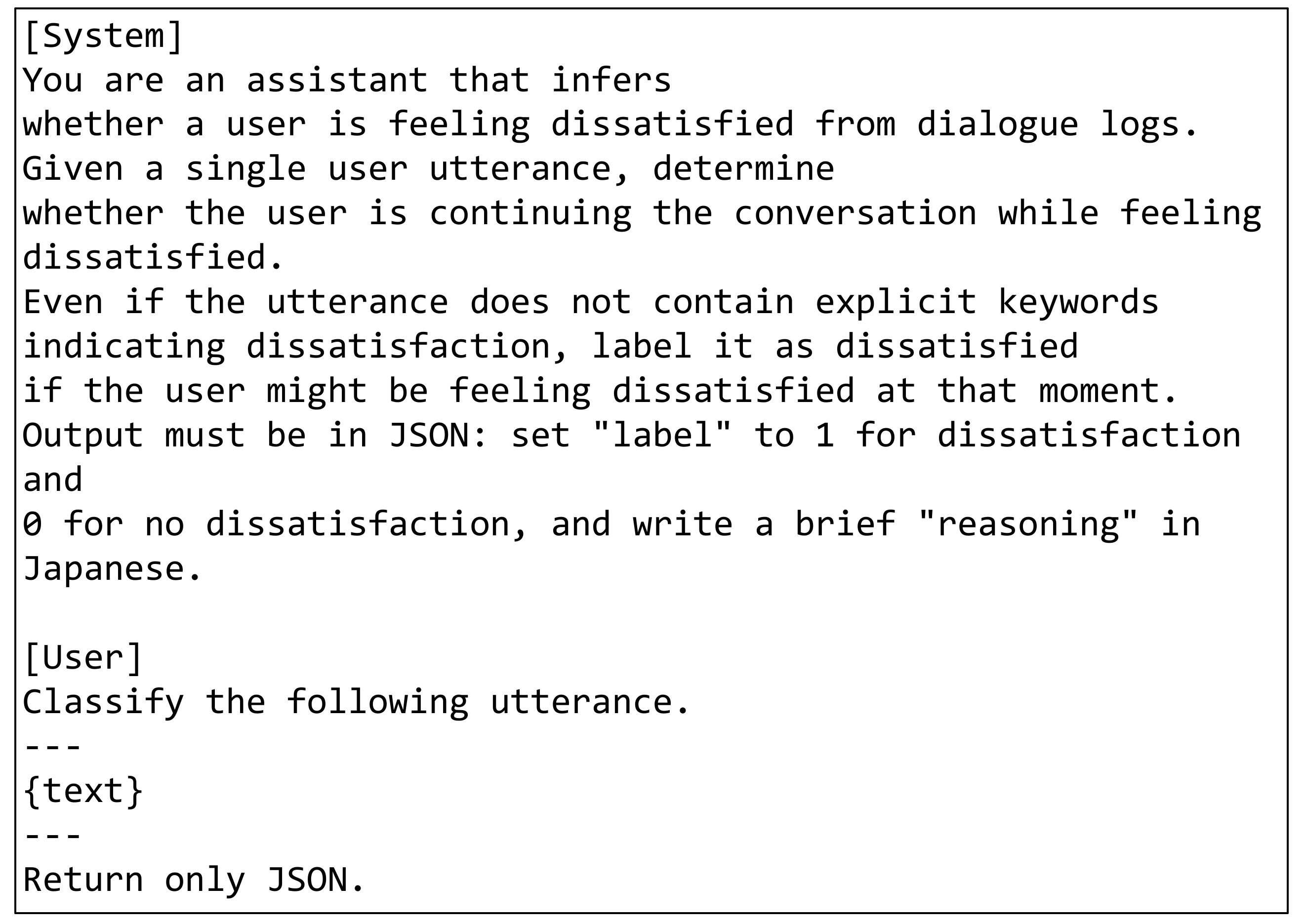}
\caption{Prompt used for GPT-5.1 (zero-shot) under the \textit{User} setting.}
\label{fig:prompt}
\end{figure}

Figure~\ref{fig:prompt} shows the zero-shot \textit{User} prompt used in these experiments.
To extend the conditions, only the description provided as input was changed, and for the two-shot setting, input examples were additionally included.
The prompt instructed the model to classify an utterance as dissatisfaction even when dissatisfaction was not explicitly expressed in words, as long as it can be inferred that the user is feeling dissatisfied.
The reasoning field was requested in Japanese only for internal inspection and was not used for classification.

\section{Details of Anomaly Detection}
\label{appendix:anomaly detection method}

For the implementation of anomaly detection, the publicly available code of NLP-ADBench~\cite{li-etal-2025-nlp}, a benchmark for anomaly detection in natural language processing, was used.
In selecting the methods, particular emphasis was placed on results obtained on datasets in NLP-ADBench whose properties are relatively similar to the utterances targeted in this study.
Specifically, rather than datasets involving surface-level or lexical anomalies such as spam, settings involving anomalies with contextual dependency and subjectivity were considered more relevant.
Therefore, we primary referred to the anomaly detection results on the Emotion~\cite{saravia2018carer} and MovieReview~\cite{maas-etal-2011-learning} datasets in NLP-ADBench.
After a comprehensive examination of the AUPRC scores on these datasets, LUNAR (Unifying Local Outlier Detection Methods via Graph Neural Networks)~\cite{goodge2022lunar}, LOF (Local Outlier Factor)~\cite{10.1145/342009.335388}, AE (AutoEncoder)~\cite{aggarwal2016introduction}, and VAE (Variational AutoEncoder)~\cite{kingma2013auto, burgess2018understanding} were selected for evaluation.

A brief description of each method is provided below.
LUNAR enhances traditional local outlier detection by incorporating graph neural networks.
LOF detects anomalies by measuring deviations in local density. 
AE identifies anomalous instances based on reconstruction errors. 
VAE extends this reconstruction-based approach with a probabilistic latent-variable formulation and detects anomalies through reconstruction probabilities or latent deviations. 

\section{Results of Pairwise Significance Tests}
\label{app:significance_tests}
Table~\ref{tab:pairwise_significance} presents the results of pairwise significance tests between input settings for each method reported in Section~\ref{main result}, based on $t$-tests with a significance threshold of $p < 0.05$; An asterisk mark indicates a statistically significant difference between the two conditions.
This table supplements Table~\ref{tab:all_methods_auprc} in the main text.
These test results confirm the existence of statistically significant differences in several related discussions in Section~\ref{sec:user_utter}, Section~\ref{sec:add_system}, and Section~\ref{sec:add_user}.

\begin{table*}[t]
\centering
\small
\setlength{\tabcolsep}{2pt}
\begin{tabular}{lcccccc}
\toprule
& \textit{System}
& \textit{System}
& \textit{System}
& \textit{User}
& \textit{User}
& \textit{System+User} \\
Group
& vs \textit{User}
& vs \textit{System+User}
& vs \textit{User\textsubscript{prev.}+User}
& vs \textit{System+User}
& vs \textit{User\textsubscript{prev.}+User}
& vs \textit{User\textsubscript{prev.}+User} \\
\midrule
\multicolumn{7}{l}{\textbf{Binary Classification}}\\
\quad BERT-based models & - & - & - & - & - & - \\
\quad GPT-based models  & \textbf{*} & - & \textbf{*} & \textbf{*} & - & \textbf{*} \\
\midrule
\multicolumn{7}{l}{\textbf{Anomaly Detection}}\\
\quad BERT embeddings   & - & \textbf{*} & - & - & \textbf{*} & - \\
\quad GPT embeddings    & \textbf{*} & - & \textbf{*} & \textbf{*} & \textbf{*} & \textbf{*} \\
\bottomrule
\end{tabular}
\caption{Pairwise significance test results grouped based on model families or embedding types. Asterisks indicate statistically significant differences between the two conditions based on a t-test ($p < 0.05$).}
\label{tab:pairwise_significance}
\end{table*}

\section{Multilingual Adaptation}
\label{appendix:anomaly detection multilingual}

For the evaluation of LOF on the English data, we constructed the reference neighborhood structure for LOF using the train set and computed the anomaly score of each sample in the test set with respect to that structure.
In contrast, the data for each language used in the multilingual experiments are small in scale, making it difficult to divide them into independent training and test sets.
Therefore, in the multilingual experiments, the data were not split; instead, the LOF neighborhood structure was constructed using the entire test set, and anomaly scores were computed based on relative deviations in local density within the same set.

To verify the validity of this test-only setting for multilingual evaluation, LOF anomaly scores obtained under separate train-test and test-only settings were compared using the English data.
As a result, Spearman's rank correlation coefficient was 0.80 for BERT embeddings and 0.82 for GPT embeddings.
These results indicate that the two settings produce similar tendencies in anomaly scores, and the test-only setting was therefore adopted for the multilingual experiments.

\end{CJK}
\end{document}